\documentclass[letterpaper,10pt,conference]{ieeeconf}
\IEEEoverridecommandlockouts
\usepackage{cite}
\usepackage{amsmath,amssymb,amsfonts}
\usepackage{algorithmic}
\usepackage{graphicx}
\usepackage{textcomp}
\usepackage{xcolor}
\usepackage{booktabs}
\usepackage{multirow}
\usepackage{caption}
\usepackage[hidelinks,hyperfootnotes=false]{hyperref}
\def\BibTeX{{\rm B\kern-.05em{\sc i\kern-.025em b}\kern-.08em
    T\kern-.1667em\lower.7ex\hbox{E}\kern-.125emX}}
\begin{document}

\title{\textbf{%
\shortstack[c]{%
Proprioception-Anchored Cross-Modal Pretraining \\
for Zero-Shot Sim-to-Real Contact-Rich Assembly%
}}}

\author{%
Yuhan Wang\textsuperscript{1},
Yurou Chen\textsuperscript{2},
Hongye Jiang\textsuperscript{1},
Gaojing Zhang\textsuperscript{3},
Wenzhao Lian\textsuperscript{1,$\dagger$}
}

\hypersetup{
    pdftitle={\textbf{Proprioception-Anchored Cross-Modal Pretraining for Zero-Shot Sim-to-Real Contact-Rich Assembly}},
    pdfauthor={Yuhan Wang, Yurou Chen, Hongye Jiang, Gaojing Zhang, Wenzhao Lian},
    pdfkeywords={contact-rich assembly, multimodal representation learning, reinforcement learning, sim-to-real transfer}
}

\twocolumn[{%
\renewcommand\twocolumn[1][]{#1}%
\maketitle
\begin{center}
    \captionsetup{type=figure}
    \includegraphics[width=1.0\textwidth]{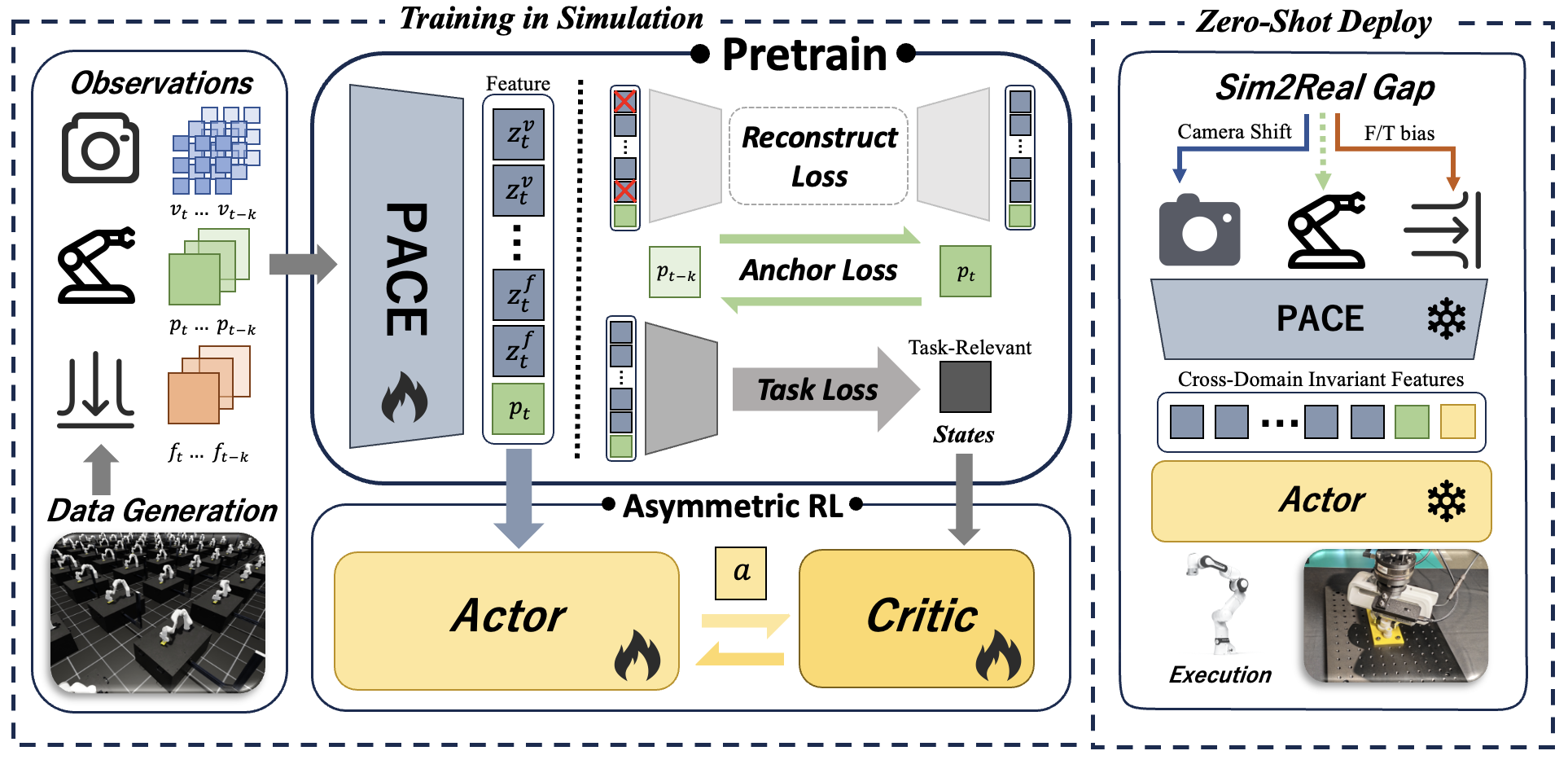}
    \captionof{figure}{Overview of the \textbf{PACE} pipeline. A privileged-state expert collects simulated demonstrations containing images, force/torque (F/T) measurements, proprioception, and privileged states. The multimodal encoder is pretrained through proprioceptive anchoring and auxiliary objectives. An asymmetric actor--critic policy is then trained on frozen encoder features and deployed directly on a real robot.}
    \label{fig:overview}
\end{center}
}]%

\begingroup
\renewcommand{\thefootnote}{}
\footnotetext{\textsuperscript{1}\,Shanghai Jiao Tong University. 
\textsuperscript{2}\,CoreNetic.ai.
\textsuperscript{3}\,University of Sussex. 
\textsuperscript{$\dagger$}\,Corresponding author.}
\endgroup

\begin{abstract}
Contact-rich assembly remains challenging because it requires submillimeter spatial accuracy and reliable interpretation of forces during sustained contact. Although simulation-based reinforcement learning offers a scalable training paradigm, discrepancies in visual observations, contact dynamics, and force/torque (F/T) measurements often limit policy transfer. We observe that proprioception is comparatively consistent across domains because joint positions are expressed in a shared calibrated coordinate system and joint velocities are computed consistently in simulation and on hardware. Based on this observation, we present PACE (Proprioception-Anchored Cross-Modal Encoder), which supervises temporal visual and F/T representations by predicting proprioceptive state transitions. Static domain-specific factors, including lighting, texture, and sensor bias, contain little information about joint motion; optimizing the proposed objective therefore suppresses their influence on the learned representation while retaining task-relevant motion cues. Policies trained on frozen PACE features are directly deployed on hardware without real-world fine-tuning or object-pose tracking. Across four contact-rich assembly tasks, PACE attains an average real-world success rate of 93.3\% and only a 2.7-percentage-point sim-to-real drop, while remaining robust to perturbations that substantially degrade pose-based and learned-fusion baselines.
\end{abstract}

\begin{keywords}
contact-rich assembly, multimodal representation learning, reinforcement learning, sim-to-real transfer
\end{keywords}

\section{Introduction}
\label{sec:intro}

Contact-rich assembly remains a fundamental challenge in robot learning. Unlike pick-and-place or tabletop rearrangement, assembly often involves submillimeter clearances, so even small misalignments can cause jamming or failure. Vision alone is therefore insufficient in many settings: successful compliant alignment requires both fine-grained spatial information and accurate interpretation of contact forces.

Simulation-based reinforcement learning (RL) provides a scalable approach to acquiring these skills, but policies trained with privileged simulation states must ultimately act from deployable observations. Common solutions replace privileged object poses with estimates from an external perception module~\cite{tang2023industreal,tang2024automate}, however, creates a brittle perception--control interface: millimeter-scale errors caused by lighting variation or calibration drift can already exceed assembly tolerances.

Common strategies for reducing this gap include domain randomization~\cite{tobin2017dr,openai2019dr}, domain adaptation~\cite{tzeng2017adversarial,bousmalis2018using}, and system identification~\cite{chebotar2019sysid}. However, domain adaptation requires real-robot data, while sufficiently randomizing or identifying visual appearance, contact dynamics, and F/T sensing remains difficult at submillimeter assembly tolerances. Policies that rely directly on these signals may therefore fail on hardware despite careful calibration of robot-side sensing and control.

Our key insight is that sensing modalities need not be equally transferable. Visual appearance and F/T measurements depend strongly on scene and contact conditions, whereas calibrated proprioception retains comparatively consistent robot-centric semantics across simulation and hardware despite residual noise and latency. We therefore use proprioceptive state transitions as a \textbf{cross-domain anchor}: temporal visual and F/T features are trained to predict these transitions, thus preserving physical motion while suppressing domain-specific nuisance factors.

We instantiate this principle as \textbf{PACE}, a \textbf{P}roprioception-\textbf{A}nchored \textbf{C}ross-modal \textbf{E}ncoder. Each modality-specific stream applies a learnable, per-dimension, zero-sum temporal kernel initialized to compute a first-order difference. The kernel suppresses static offsets and extracts a compact motion code. During pretraining, a cycle-consistency objective anchors these codes to proprioception, while auxiliary objectives preserve task-relevant spatial and contact information. We then freeze the encoder, train an RL policy on its features, and deploy the resulting policy on hardware without adaptation.

We evaluate PACE on four contact-rich assembly tasks with distinct challenges: PegInsert and GearMesh from the Factory benchmark, and JointAssemble (a square plug with rotational constraints) and KnobTighten (a knob-and-socket task with sloped contact geometry and increased reliance on F/T feedback) from the AutoMate dataset. Across these tasks, PACE achieves a 93.3\% average real-world success rate, with an average sim-to-real decrease of only 2.7 percentage points. In contrast, pose-based and learned-fusion baselines exhibit substantially larger decreases on hardware.

The main contributions of this work are as follows:
\begin{itemize}
    \item We introduce \textbf{proprioceptive anchoring}, which uses comparatively transferable robot-centric proprioception to supervise domain-sensitive visual and F/T representations.
    \item We instantiate this principle in \textbf{PACE}, combining learnable zero-sum temporal differencing, bidirectional proprioceptive-transition prediction, and auxiliary objectives that preserve task-relevant spatial and contact information.
    \item Across four contact-rich assembly tasks, PACE achieves a 93.3\% average hardware success rate with an average sim-to-real decrease of 2.7 percentage points; controlled ablations identify proprioceptive anchoring as the main contributor to transfer, while real-world perturbation tests demonstrate robustness to visual and F/T shifts.
\end{itemize}
\section{Related Work}
\label{sec:related}

\subsection{Contact-Rich Robotic Assembly}

Classical assembly methods rely on engineered compliance~\cite{whitney1982quasi,mason1981mechanics}, hybrid force--position control~\cite{raibert1981hybrid}, or impedance control~\cite{hogan1985impedance}. Although effective in structured settings, these approaches often require task- and geometry-specific design. Reinforcement learning has enabled contact-rich manipulation in simulation and on hardware~\cite{levine2016endtoend,lillicrap2016continuous,kalashnikov2018qtopt}. The Factory benchmark~\cite{narang2022factory} provides standardized assembly tasks in Isaac Gym, with privileged simulation states used for policy learning, while AutoMate~\cite{tang2024automate} scales simulation-based training to 100 assembly geometries.

Hardware-oriented learning systems variously rely on vision, force feedback, or data collected on real robots. Methods guided by perception estimate object poses before executing policies trained in simulation. IndustReal~\cite{tang2023industreal}, for example, combines Mask R-CNN detection with camera calibration using AprilTags, while AutoMate also uses pose estimates from perception during deployment. Such modular interfaces can be sensitive to estimation errors under tight assembly clearances. Methods guided by force feedback, including FORGE~\cite{noseworthy2025forge} and CoRMA~\cite{wang2026corma}, improve contact exploration but do not use vision for initial spatial alignment. ResiP~\cite{ankile2024resip} trains a corrective residual policy in simulation on top of a frozen behavior-cloned base policy; its deployable visual policy is then distilled from simulated rollouts and co-trained with a small set of real demonstrations. These approaches differ in their reliance on external pose estimates, available sensing, and robot data. PACE instead learns deployable visual and F/T representations entirely from simulated data, enabling direct policy transfer from onboard observations without an external pose estimator or object tracker.

\subsection{Sim-to-Real Transfer}

Bridging the sim-to-real gap remains a central challenge in robot learning. Common strategies include domain randomization~\cite{tobin2017dr,openai2019dr,peng2018sim2real,mehta2020activedr}, which samples visual and physical parameters during training; domain adaptation~\cite{tzeng2017adversarial,bousmalis2018using}, which aligns feature distributions using real-robot data; and system identification~\cite{chebotar2019sysid}, which calibrates simulation parameters against hardware. These strategies improve transfer by broadening the training distribution, aligning observations across domains, or reducing discrepancies in simulated dynamics. However, residual errors, real-data requirements, and task-specific calibration remain problematic at submillimeter assembly tolerances.

Asymmetric training gives a critic or teacher privileged information while restricting the actor or student to deployable observations~\cite{pinto2018asymmetric,kumar2021rma}. This separation allows privileged simulation states to guide policy learning without requiring privileged information during deployment, but does not ensure cross-domain consistency of the actor's representations. PACE retains domain-randomized asymmetric policy learning but focuses on representation transfer when sensing modalities exhibit different degrees of cross-domain consistency. Rather than aligning simulated features with real data, PACE uses comparatively consistent proprioceptive transitions to supervise temporal visual and F/T representations during simulation-only pretraining.

\subsection{Representation Learning for Robotic Manipulation}

General-purpose visual encoders, including R3M~\cite{r3m}, VIP~\cite{vip}, and VC-1~\cite{vc1}, learn reusable representations from large-scale datasets but are not designed specifically for the fine-grained spatial reasoning required in submillimeter assembly. DINOv2~\cite{dinov2} is a self-supervised vision transformer~\cite{dosovitskiy2021vit} that exhibits spatial correspondence and serves as our frozen visual backbone. Relevant self-supervised objectives include masked autoencoding~\cite{mae} and contrastive representation learning~\cite{chen2020simclr,grill2020byol,he2020moco}.

Multimodal policies combine sensing streams through feature concatenation~\cite{lee2019tactile,guzey2023visiontouch} or cross-attention~\cite{diaz2025auginsert}, but fusion alone does not enforce cross-domain consistency. Tactile-only methods~\cite{dong2021tactile,kim2022activecontact} support insertion through contact sensing but provide limited information for initial alignment.

Dynamics-aware representation learning uses forward and inverse objectives to retain control-relevant information. DynaMo~\cite{dynamo} applies such objectives to images, while AFRO~\cite{liang2026afro} extends feature differencing and inverse consistency to 3-D point clouds. MSDP~\cite{krohn2026msdp} learns visual, F/T, and proprioceptive representations through masked autoencoding and cross-sensor prediction. Proprioception provides a reliable signal beyond assembly: KiVi~\cite{li2026kivi} supports locomotion under visual degradation, while TNavRL~\cite{huang2026tnavrl} relates noisy vision to proprioceptive states for zero-shot navigation. Large vision--language--action models~\cite{brohan2023rt2,padalkar2023openx} and diffusion policies~\cite{chi2023diffusion} instead emphasize general-purpose policy learning.

PACE grounds visual and F/T motion representations in proprioceptive transitions, using robot motion as their shared physical reference. Zero-sum temporal kernels keep temporal changes while suppressing static offsets; spatial-alignment and engagement objectives retain task-relevant information; and masked reconstruction with stop-gradient targets promotes cross-stream information sharing and robustness to partial feature loss. Together, these components target transferable motion and contact representations for zero-shot sim-to-real assembly.
\section{Problem Formulation}
\label{sec:problem}

We consider \textbf{quasi-static contact-rich assembly}. Given a grasped object and a target receptacle, the robot must complete the assembly within a task-specific positional tolerance $\delta \in [0.1,1.0]$~mm. At each time step $t$, the observations comprise images $\mathbf{I}^L_t$ and $\mathbf{I}^R_t$ from two wrist-mounted cameras, a six-axis F/T measurement $\mathbf{F}_t$, joint positions $\mathbf{q}_t \in \mathbb{R}^7$, joint velocities $\dot{\mathbf{q}}_t \in \mathbb{R}^7$, and the previous action $\mathbf{a}_{t-1} \in \mathbb{R}^6$. The policy outputs an action $\mathbf{a}_t \in \mathbb{R}^6$ specifying an end-effector pose increment $(\Delta x, \Delta y, \Delta z, \Delta\theta_{\mathrm{roll}}, \Delta\theta_{\mathrm{pitch}}, \Delta\theta_{\mathrm{yaw}})$ under impedance control. The controller holds roll and pitch fixed to maintain an upright gripper, so only the position and yaw increments affect the executed motion.

Let $\mathbf{s}_t$ denote the latent physical state of the assembly system, comprising the robot configuration and velocity, the poses and velocities of the manipulated part and receptacle, and the contact configuration. We consider three sources of the sim-to-real gap:
\begin{enumerate}
    \item \textbf{Visual observation gap}: $P_{\mathrm{sim}}(\mathbf{I}_t\mid\mathbf{s}_t) \neq P_{\mathrm{real}}(\mathbf{I}_t\mid\mathbf{s}_t)$, owing to differences between rendered and real camera observations.
    \item \textbf{F/T sensing gap}: $P_{\mathrm{sim}}(\mathbf{F}_t\mid\mathbf{s}_t) \neq P_{\mathrm{real}}(\mathbf{F}_t\mid\mathbf{s}_t)$, owing to differences between simulated and real F/T sensor characteristics.
    \item \textbf{Physical interaction gap}: $P_{\mathrm{sim}}(\mathbf{s}_{t+1}\mid\mathbf{s}_t,\mathbf{a}_t) \neq P_{\mathrm{real}}(\mathbf{s}_{t+1}\mid\mathbf{s}_t,\mathbf{a}_t)$, owing to differences in contact mechanics and robot--environment interactions.
\end{enumerate}
Domain randomization is used during large-scale RL training to mitigate the physical interaction gap.

We define the proprioceptive observation as
$\mathbf{p}_t = [\mathbf{q}_t;\dot{\mathbf{q}}_t] \in \mathbb{R}^{14}$.
In contrast to visual and F/T measurements, proprioception is represented in
a common joint coordinate system in simulation and on hardware. We therefore
model the proprioceptive observation distributions as approximately aligned:
\begin{equation}
P_{\mathrm{sim}}(\mathbf{p}_t \mid \mathbf{s}_t)
\approx
P_{\mathrm{real}}(\mathbf{p}_t \mid \mathbf{s}_t).
\end{equation}
Hardware joint positions are obtained from calibrated joint encoders, and joint velocities in both domains are computed by finite differencing joint positions at the control frequency. The shared joint coordinate system and consistent velocity computation provide a common cross-domain reference despite residual noise, latency, and quantization. We therefore use proprioception as a cross-domain anchor.

Our objective is to learn an encoder $\phi:(\mathbf{I}^L_t,\mathbf{I}^R_t,\mathbf{F}_t,\mathbf{p}_t) \mapsto \mathbf{z}_t \in \mathbb{R}^{256}$, where $\mathbf{z}_t=[\mathbf{z}^{v}_t;\mathbf{z}^{f}_t]$ retains task-relevant information while suppressing domain-specific nuisance factors. We then train a policy $\pi(\mathbf{a}_t \mid \mathbf{z}_t,\mathbf{p}_t,\mathbf{a}_{t-1})$ in simulation and evaluate it on hardware. Throughout this paper, zero-shot sim-to-real transfer refers to directly deploying a simulation-trained policy on hardware without real-world policy adaptation.


\begin{figure*}[t]
\centering
\includegraphics[width=1.0\textwidth]{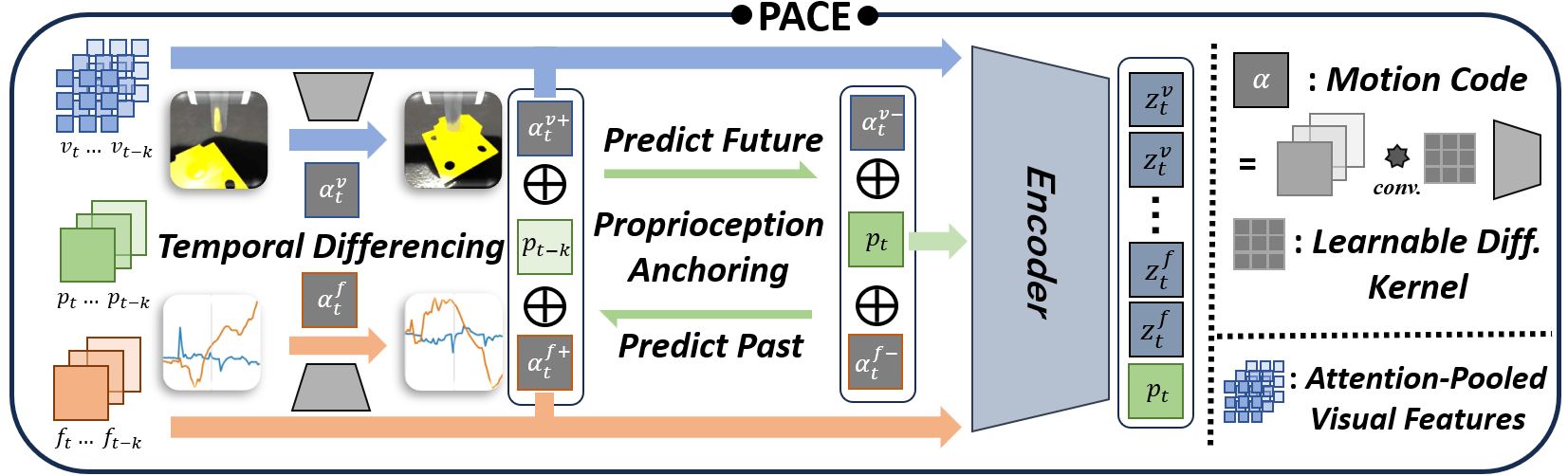}
\caption{PACE encoder architecture. A frozen visual backbone followed by attention pooling extracts per-frame visual features, while an F/T multilayer perceptron (MLP) extracts per-frame force features. Stream-specific temporal kernels aggregate the preceding \(T\) features into forward motion codes \(\boldsymbol{\alpha}^{v+}_{t}\) and \(\boldsymbol{\alpha}^{f+}_{t}\); negated inputs produce backward codes \(\boldsymbol{\alpha}^{v-}_{t}\) and \(\boldsymbol{\alpha}^{f-}_{t}\). Each stream combines its current feature, motion code, and proprioception \(\mathbf p_t\) to produce a latent representation.}
\label{fig:arch}
\end{figure*}

\section{Method}
\label{sec:method}

PACE consists of two stages. First, we pretrain an encoder to predict domain-consistent physical quantities, thereby reducing the sensitivity of its features to domain-specific observations. Second, we freeze the encoder, train an RL policy on the resulting features, and deploy the policy directly on hardware. Fig.~\ref{fig:overview} summarizes the complete pipeline, including expert data collection, encoder pretraining, policy learning, and deployment, while Fig.~\ref{fig:arch} details the encoder architecture.

\subsection{Expert Demonstrations}

For each task, we train a privileged-state PPO policy~\cite{schulman2017ppo} in Isaac Lab~\cite{mittal2025isaaclab}. The policy observes ground-truth poses, velocities, and contact states, and we train it using the default Factory domain-randomization settings~\cite{narang2022factory} until convergence. We roll out the stochastic expert to collect 300 trajectories per task. At each time step, we record images from both wrist cameras, a short history of wrist F/T measurements, joint positions and velocities, the executed action, and a binary engagement label obtained from simulated contact detection. We partition the resulting dataset into training and validation sets using an 80:20 split.

\subsection{Encoder Architecture}

Each modality follows the same temporal-differencing pipeline: encoding each frame, buffering the preceding \(T\) features, extracting a motion code with a learnable temporal kernel, and fusing the result with proprioception to produce the stream output. The temporal kernel is a per-dimension depthwise 1-D convolution initialized to approximate a first-order difference, $[-1,1,0,\ldots]$. On every forward pass, we subtract the per-channel mean of the weights to enforce an exactly zero-sum kernel. This constraint removes constant offsets, such as static appearance components or sensor bias, from the motion code and retains temporal changes in the visual and force signals.

For the visual stream, a frozen backbone encodes images from both wrist cameras. Learned attention pooling and cross-view fusion then produce the per-frame feature $\mathbf{v}_t$. From the preceding $T$ visual features, the temporal kernel extracts a forward motion code $\boldsymbol{\alpha}^{v+}_t$ that represents object motion, gripper approach, and contact-induced deflection. The stream output $\mathbf{z}^{v}_t$ combines $\mathbf{v}_t$, $\boldsymbol{\alpha}^{v+}_t$, and proprioception $\mathbf{p}_t$; absolute appearance is excluded from the motion pathway. For the force stream, a multilayer perceptron (MLP) encodes a recent window of six-axis F/T measurements into a compact feature $\mathbf{f}_t$, from which a separate temporal kernel extracts the force-motion code $\boldsymbol{\alpha}^{f+}_t$. The output $\mathbf{z}^{f}_t$ combines $\mathbf{f}_t$, $\boldsymbol{\alpha}^{f+}_t$, and $\mathbf{p}_t$, suppressing static sensor bias in the motion code while retaining contact transients.

\subsection{Proprioception-Anchored Cross-Modal Pretraining}

We pretrain the encoder on the collected demonstrations using proprioceptive anchoring, task-related supervision, and cross-modal reconstruction. The visual backbone remains frozen throughout pretraining.

\textbf{Proprioceptive Anchoring.} This objective promotes cross-domain invariance. Given the motion codes $(\boldsymbol{\alpha}^{v+}_t, \boldsymbol{\alpha}^{f+}_t)$ and the proprioceptive state $K$ steps earlier, a lightweight MLP head predicts the proprioceptive innovation $\Delta\mathbf{p}_t = \mathbf{p}_t - \mathbf{p}_{t-K}$ and adds it to the earlier state. Predicting the innovation rather than the absolute state precludes the trivial solution of copying $\mathbf{p}_{t-K}$, which would leave the motion codes unconstrained. Cycle consistency is imposed by re-extracting motion codes from negated temporal differences and using the same head to predict the earlier state:

\begin{equation}
\boldsymbol{\alpha}^{v+}_t = g_v(\Delta\mathbf{v}_t),\;\;
\boldsymbol{\alpha}^{f+}_t = g_f(\Delta\mathbf{f}_t),
\end{equation}

\begin{equation}
\hat{\mathbf{p}}_t = \mathbf{p}_{t-K} + \mathrm{MLP}(\mathbf{p}_{t-K} \oplus \boldsymbol{\alpha}^{v+}_t \oplus \boldsymbol{\alpha}^{f+}_t),
\end{equation}

\begin{equation}
\hat{\mathbf{p}}_{t-K} = \hat{\mathbf{p}}_t + \mathrm{MLP}(\hat{\mathbf{p}}_t \oplus \boldsymbol{\alpha}^{v-}_t \oplus \boldsymbol{\alpha}^{f-}_t).
\end{equation}

The proprioceptive anchoring loss $\mathcal{L}_{\mathrm{anchor}}$ penalizes the smooth-$L_1$ error in both directions, with the backward branch imposing cycle consistency on the learned motion representation. The two directions share a single prediction head, and gradients through the forward prediction are stopped before the backward prediction is computed. Backward motion codes are recomputed from negated temporal differences, e.g., $\boldsymbol{\alpha}^{v-}_t = g_v(-\Delta\mathbf{v}_t)$. Domain-specific appearance and static sensor offsets provide little information about proprioceptive transitions. Minimizing the loss therefore makes the temporal representations encode the physical motion associated with these transitions.

\textbf{Task-Related Supervision.} We train two prediction heads using state labels that are readily available in simulation. Conditioned on the visual representation and previous action, the spatial alignment head predicts the normalized planar vector from the end effector to the target keypoint; for JointAssemble, it additionally regresses the yaw error. The spatial alignment loss $\mathcal{L}_{\mathrm{align}}$ uses a cosine loss on the planar direction whenever the target displacement has a well-defined direction and a smooth-$L_1$ loss on yaw when applicable. The engagement head takes both stream representations as input and predicts whether the held part is engaged with the fixed receptacle; it is trained using the binary cross-entropy loss $\mathcal{L}_{\mathrm{eng}}$. These objectives provide direct supervision for spatial alignment and contact engagement, respectively.

\textbf{Cross-Modal Reconstruction.} To improve robustness to partial feature corruption and facilitate information sharing across sensing streams, we apply masked reconstruction in feature space. We uniformly mask half of the dimensions in the concatenated vector $[\mathbf{z}^{v}_t, \mathbf{z}^{f}_t, \mathbf{p}_t]$ and use a shallow MLP to reconstruct the complete vector. The reconstruction loss includes the visible dimensions, and the target is detached from the computation graph. Gradients therefore pass through the reconstructor to the visual and force pathways but not through the target. Locally correlated dimensions can be recovered within a modality, whereas the remaining information must be inferred across modalities. This reconstruction couples the two pathways through a shared multimodal context; its contribution is evaluated by the `w/o Masked Reconstruction' ablation in Table~\ref{tab:sim2real}.

\textbf{Pretraining Procedure.} We minimize $\mathcal{L}=\lambda_{\mathrm{anchor}}\mathcal{L}_{\mathrm{anchor}}+\lambda_{\mathrm{align}}\mathcal{L}_{\mathrm{align}}+\lambda_{\mathrm{eng}}\mathcal{L}_{\mathrm{eng}}+\lambda_{\mathrm{recon}}\mathcal{L}_{\mathrm{recon}}$, with $(\lambda_{\mathrm{anchor}},\lambda_{\mathrm{align}},\lambda_{\mathrm{eng}},\lambda_{\mathrm{recon}})=(1,2,2,1)$. We apply color jitter, blur, and cropping to the images and add annealed Gaussian noise to the F/T measurements to approximate real sensor noise. The encoder is trained for up to 100 epochs with early stopping. 

\subsection{Policy Training and Deployment}

We freeze the pretrained encoder and use its features as observations for RL policy training. The encoder runs online at every simulation step and maintains the same temporal feature buffers used during pretraining. We employ an asymmetric actor--critic architecture: the actor observes only the latent features, proprioception, and the previous action, whereas the critic receives privileged simulation state for value estimation. Thus, the actor depends only on observations available during deployment. We train all RL policies for 200 iterations in Isaac Lab using the same policy architecture, hyperparameters, and domain-randomization settings, thereby isolating the effect of the observation representation.

For deployment, we transfer the frozen encoder and policy to the robot without parameter updates or real-world training data. Camera images and F/T measurements undergo the same preprocessing used in simulation, and the encoder updates its temporal buffers online. The policy commands incremental end-effector poses through Cartesian impedance control~\cite{hogan1985impedance}. Before evaluation, we perform a one-time hardware configuration shared across tasks: we fix the wrist-camera mounts and exposure settings, the end-effector frame offset, and the impedance gains of the robot; we also zero the F/T sensor. No policy or encoder parameters are updated on real-world data.

\section{Experimental Setup}
\label{sec:setup}

\subsection{Tasks and Evaluation}

We evaluate PACE on four contact-rich assembly tasks that span distinct geometric and contact-sensing challenges. Two tasks are adapted from the Factory benchmark~\cite{narang2022factory}. \textbf{PegInsert} requires insertion of a 16-mm peg into a receptacle with approximately 0.1-mm radial clearance and primarily tests fine visual alignment. \textbf{GearMesh} requires a 30-mm-diameter gear to mesh with adjacent fixed gears and tests contact-sensitive translational alignment.

\begin{figure}[t]
\centering
\includegraphics[width=0.47\textwidth]{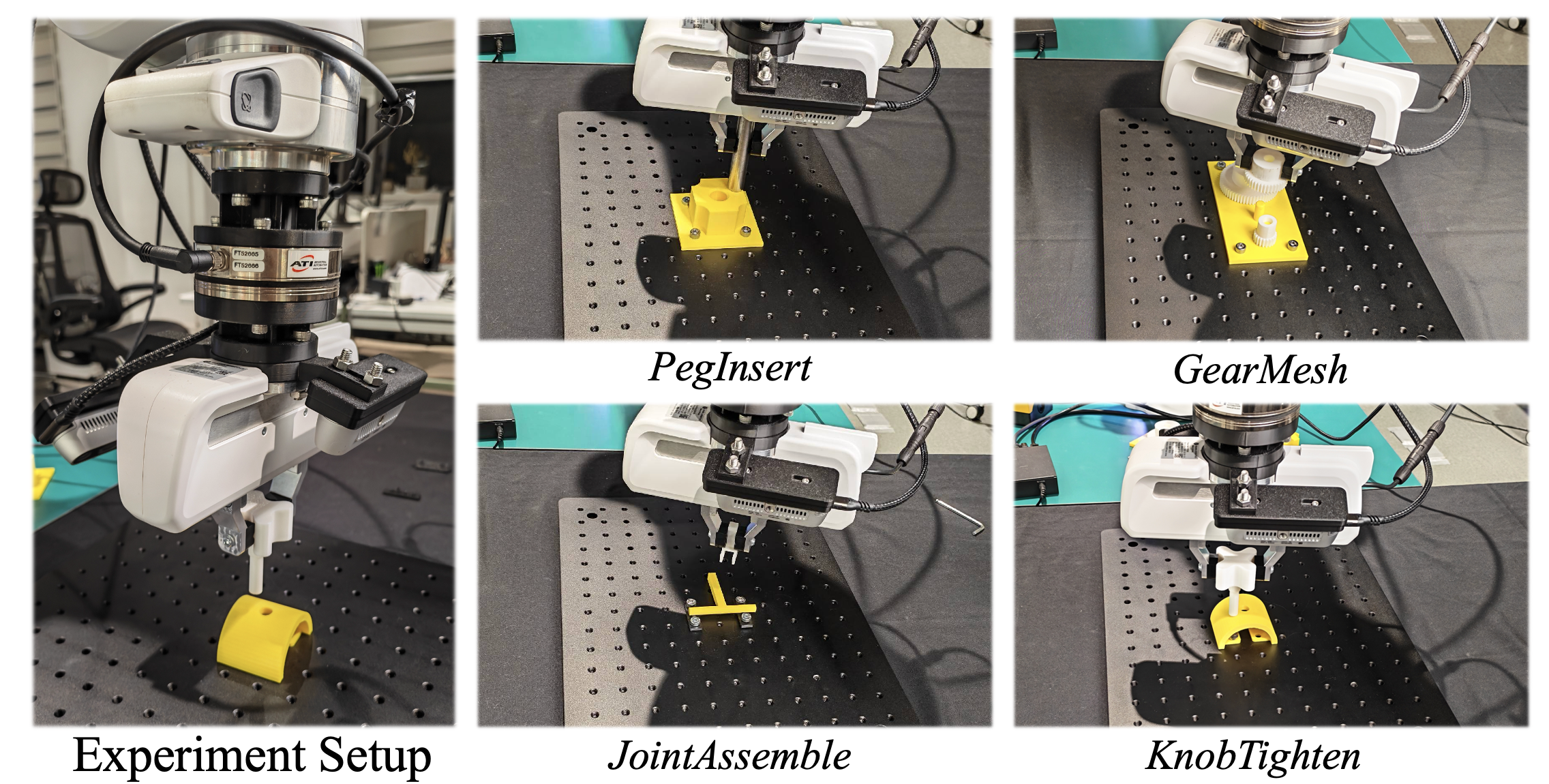}
\caption{Robot setup and representative scenes for the four assembly tasks.}
\label{fig:tasks}
\end{figure}

The other two tasks are drawn from the AutoMate dataset~\cite{tang2024automate}. \textbf{JointAssemble} (ID 00471) uses a $2{\times}1{\times}3.3$-cm square plug that requires both translational and yaw alignment during insertion. Its auxiliary spatial alignment head therefore predicts planar direction and yaw error. \textbf{KnobTighten} (ID 00855) uses a $3.7{\times}3.7{\times}6.5$-cm knob and a sloped socket whose contact geometry increases the importance of F/T feedback during engagement. Fig.~\ref{fig:tasks} shows the robot setup and representative scenes for all four tasks.

All experiments use a Franka FR3 manipulator equipped with two $224\times224$-pixel wrist-mounted RGB cameras~\cite{spector2021insertionnet} and a six-axis wrist F/T sensor. At 15~Hz frequency, the policy outputs 6-D pose increments; the controller fixes roll and pitch, leaving 4 controllable DoF: translation and yaw. An episode is counted as successful if the assembly reaches the target pose within the task-specific tolerance.

For each of three random seeds, we evaluate the resulting policy over 100 simulated episodes under domain randomization and report the mean success rate and standard deviation across seeds. For hardware evaluation, we select, for each method--task pair, the checkpoint with the highest simulated success rate among the three seeds and deploy it for 30 consecutive trials. We randomize the initial gripper and target poses before every trial and determine success from the final pose and contact conditions. The same checkpoint-selection and evaluation protocol is applied to all methods.

\subsection{Baselines}

We compare PACE against three baselines representing privileged-state control, direct policy learning from unadapted sensor features, and learned multimodal fusion.

\textbf{Pose-Based PPO.} This baseline uses the privileged-state PPO policy described in Sec.~\ref{sec:method}. In simulation, the policy receives ground-truth object poses, keypoint relations, and contact variables. During deployment, the policy weights remain fixed, and unavailable geometric quantities are replaced with estimates from an external perception module following IndustReal~\cite{tang2023industreal}. Object detection and AprilTag-based camera calibration provide planar part poses $(x,y,\theta)$. This baseline represents a conventional modular deployment in which errors at the perception--control interface contribute to the sim-to-real performance decrease.

\textbf{No-Pretrain RL.} The actor receives the concatenation of frozen DINOv2 visual features, the F/T history, proprioception, and the previous action. This baseline omits task-specific encoder pretraining, proprioceptive anchoring, and masked reconstruction. We initialize the policy randomly and train it with PPO using the same RL budget as PACE. The comparison evaluates whether generic visual features and raw multimodal observations are sufficient to learn transferable assembly policies.

\textbf{AugInsert.} We implement AugInsert using a Perceiver IO architecture~\cite{jaegle2021perceiver}. DINOv2 patch tokens and F/T embeddings receive modality-specific positional encodings and are fused through cross-attention~\cite{diaz2025auginsert}. The model is first trained via behavior cloning using the expert demonstrations collected for PACE and is then fine-tuned with PPO under the same RL budget. PACE and AugInsert use the same expert demonstrations for encoder pretraining and behavior-cloning initialization, respectively. All policies use the same PPO budget and real-robot evaluation protocol. This baseline evaluates whether behavior-cloning initialization and cross-attention-based multimodal fusion are sufficient for robust sim-to-real transfer.

\subsection{Ablations}

We evaluate four ablations to quantify the contributions of proprioceptive anchoring, force information, visual information, and masked latent reconstruction. Unless otherwise stated, all variants retain the full-model loss weights and training settings. For the single-stream variants, each objective uses the available representation, and $\mathcal{L}_{\mathrm{align}}$ is omitted when the visual stream is removed.

\textbf{Without Proprioceptive Anchoring ($\mathcal{L}_{\mathrm{anchor}}$).} This variant sets the weight of $\mathcal{L}_{\mathrm{anchor}}$ to zero while retaining the spatial alignment, engagement, and reconstruction objectives. It measures the contribution of proprioceptive-transition prediction to cross-domain transfer.

\textbf{Without the Force Stream.} We remove the force stream from both encoder pretraining and policy input, such that the policy observes $[\mathbf{z}^{v}_t,\mathbf{p}_t,\mathbf{a}_{t-1}]$. This variant measures the contribution of F/T information, particularly in tasks involving meshing, rotational mismatch, or force-sensitive engagement with sloped geometry.

\textbf{Without the Visual Stream.} We remove the visual stream from both encoder pretraining and policy input, such that the policy observes $[\mathbf{z}^{f}_t,\mathbf{p}_t,\mathbf{a}_{t-1}]$. This variant evaluates whether force and proprioception alone provide sufficient spatial information for initial alignment.

\textbf{Without Masked Reconstruction ($\mathcal{L}_{\mathrm{recon}}$).} This variant sets the weight of $\mathcal{L}_{\mathrm{recon}}$ to zero while retaining proprioceptive anchoring and privileged auxiliary supervision. It measures the effect of masked latent reconstruction on nominal transfer and robustness to real-world perturbations.

\section{Results and Analysis}
\label{sec:results}

\subsection{Zero-Shot Sim-to-Real Transfer}

\begin{figure}[t]
\centering
\includegraphics[width=\columnwidth]{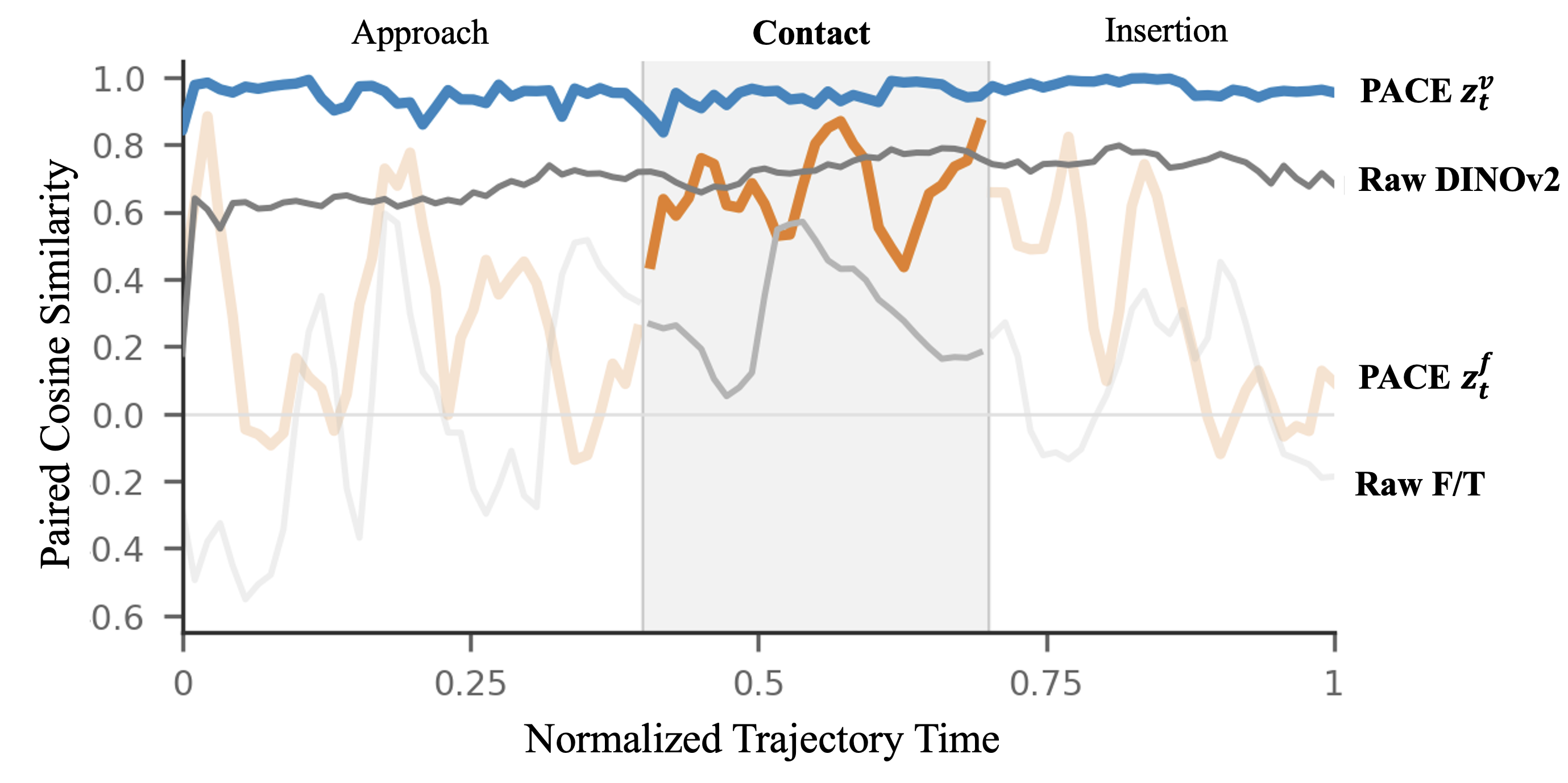}
\caption{Cross-domain feature similarity along a representative pair of matched simulation and hardware trajectories. PACE features exhibit higher cosine similarity than the corresponding unadapted representations, particularly for F/T signals during contact.}
\label{fig:domaingap}
\end{figure}

\begin{table*}[t]
\centering
\caption{Zero-shot sim-to-real transfer. Sim: mean success rate (\%) $\pm$ standard deviation over three seeds (100 episodes per seed). Real: success rate (\%) over 30 hardware trials using the best-simulation checkpoint for each method--task pair. $\Delta=\mathrm{Sim}-\mathrm{Real}$ (pp; lower is better). The 95\% Wilson interval half-widths for real-world rates are 8--17~pp. ``---'' denotes no hardware evaluation.}
\label{tab:sim2real}
\footnotesize
\setlength{\tabcolsep}{3.6pt}
\begin{tabular}{@{}lcccccccccccccc@{}}
\toprule
& \multicolumn{2}{c}{\textbf{Average}} & \multicolumn{3}{c}{\textbf{PegInsert}} & \multicolumn{3}{c}{\textbf{GearMesh}} & \multicolumn{3}{c}{\textbf{JointAsm}} & \multicolumn{3}{c}{\textbf{KnobTight}} \\
\cmidrule(lr){2-3}\cmidrule(lr){4-6}\cmidrule(lr){7-9}\cmidrule(lr){10-12}\cmidrule(lr){13-15}
& Real & $\Delta$ & Sim & Real & $\Delta$ & Sim & Real & $\Delta$ & Sim & Real & $\Delta$ & Sim & Real & $\Delta$ \\
\midrule
Pose-Based PPO      & 77.5 & 18.9 & \textbf{97.7}$_{\pm1.5}$ & 83.3 & 14.3 & \textbf{96.7}$_{\pm1.5}$ & 80.0 & 16.7 & \textbf{95.3}$_{\pm0.6}$ & 73.3 & 22.0 & \textbf{96.0}$_{\pm1.0}$ & 73.3 & 22.7 \\
AugInsert           & 68.3 & 18.4 & 92.3$_{\pm1.5}$ & 76.7 & 15.7 & 85.3$_{\pm4.5}$ & 70.0 & 15.3 & 85.3$_{\pm1.5}$ & 66.7 & 18.7 & 84.0$_{\pm3.0}$ & 60.0 & 24.0 \\
No-Pretrain RL      & 7.5 & 43.7 & 56.3$_{\pm19.6}$ & 13.3 & 43.0 & 53.3$_{\pm25.7}$ & 6.7 & 46.7 & 39.3$_{\pm16.0}$ & 6.7 & 32.7 & 56.0$_{\pm15.4}$ & 3.3 & 52.7 \\
\midrule

\textbf{PACE (Full)} & \textbf{93.3} & \textbf{2.7} & 97.3$_{\pm1.2}$ & \textbf{96.7} & \textbf{0.7} & 95.7$_{\pm1.2}$ & \textbf{93.3} & \textbf{2.3} & \textbf{95.3}$_{\pm1.5}$ & \textbf{90.0} & \textbf{5.3} & 95.7$_{\pm1.2}$ & \textbf{93.3} & \textbf{2.3} \\
\midrule
w/o $\mathcal{L}_{\mathrm{anchor}}$     & 65.8 & 25.5 & 92.7$_{\pm2.3}$ & 70.0 & 22.7 & 92.0$_{\pm3.6}$ & 53.3 & 38.7 & 89.3$_{\pm4.0}$ & 70.0 & 19.3 & 91.3$_{\pm2.5}$ & 70.0 & 21.3 \\
w/o Force Stream                   & 75.8 & 8.8 & 94.3$_{\pm2.1}$ & 86.7 & 7.7 & 84.3$_{\pm1.5}$ & 70.0 & 14.3 & 79.3$_{\pm4.0}$ & 76.7 & 2.7 & 80.7$_{\pm2.5}$ & 70.0 & 10.7 \\
w/o Masked Reconstruction          & 85.8 & 6.1 & 92.0$_{\pm2.6}$ & 86.7 & 5.3 & 91.7$_{\pm4.5}$ & 86.7 & 5.0 & 92.3$_{\pm2.1}$ & 83.3 & 9.0 & 91.7$_{\pm2.1}$ & 86.7 & 5.0 \\
w/o Visual Stream                  & --- & --- & 9.0$_{\pm5.6}$ & --- & --- & 8.7$_{\pm7.5}$ & --- & --- & 5.0$_{\pm2.6}$ & --- & --- & 9.3$_{\pm1.5}$ & --- & --- \\
\bottomrule
\end{tabular}
\end{table*}

Table~\ref{tab:sim2real} first summarizes average hardware performance and then reports task-level simulation and hardware results. PACE achieves the highest average real-world success rate, 93.3\%, and the smallest average transfer decrease, 2.7~pp. The strongest baseline, Pose-Based PPO, reaches 77.5\% average hardware success with an 18.9-pp decrease, while AugInsert reaches 68.3\% with an 18.4-pp decrease. No-Pretrain RL transfers poorly, attaining only 7.5\% average hardware success and a 43.7-pp decrease. Across individual tasks, PACE maintains 90.0--96.7\% hardware success and a transfer decrease of 0.7--5.3~pp.

Simulation results separate representation learning from cross-domain transfer. Pose-Based PPO and PACE achieve similar average simulated success rates of 96.4\% and 96.0\%, respectively; however, replacing privileged poses with estimates on hardware produces a much larger decrease for Pose-Based PPO. AugInsert reaches 86.7\% average simulated success with behavior-cloning initialization, whereas No-Pretrain RL reaches only 51.2\% and exhibits high variability across seeds. The variant without vision attains only 5--9\% simulated success, showing that F/T sensing and proprioception alone are insufficient for initial alignment. Together, these results show that generic visual features and multimodal fusion alone are insufficient to match PACE's transfer performance.

We next use paired cosine similarity as a diagnostic of cross-domain feature consistency. For each task, five simulated 7-DoF joint trajectories are replayed on the real robot at 15~Hz, and the resulting features are paired frame by frame. We average cosine similarity over the contact phase of each trajectory pair. Across the resulting 20 pairs, the PACE visual and F/T features achieve mean similarities of $0.91 \pm 0.08$ and $0.64 \pm 0.23$, compared with $0.69 \pm 0.08$ for frozen DINOv2 features and $0.23 \pm 0.20$ for unprocessed F/T measurements. The $\pm$ values denote 95\% confidence-interval half-widths. Fig.~\ref{fig:domaingap} shows one representative pair. These similarities are consistent with reduced cross-domain discrepancy in the PACE representations; they serve as a diagnostic, while Table~\ref{tab:sim2real} provides the task-level evidence for transfer.

The ablations distinguish each component's effect on simulated performance and transfer. Without proprioceptive anchoring, the average $\Delta$ rises from 2.7 to 25.5~pp, more than for any other hardware-tested ablation, whereas simulated success falls by just 4.7~pp. This disparity suggests that anchoring primarily improves cross-domain transfer rather than task performance in simulation. Removing the force pathway decreases average simulated success by 11.3~pp, with the largest task-level decreases occurring on GearMesh, JointAssemble, and KnobTighten, where contact sensing helps resolve meshing, orientation mismatch, and surface engagement. Removing masked reconstruction decreases simulated success by 4.1~pp and increases the average $\Delta$ from 2.7 to 6.1~pp. Its role under real-world disturbances is examined further in Fig.~\ref{fig:robustness}.

\subsection{Real-World Robustness Under Perturbations}

\begin{figure}[t]
\centering
\includegraphics[width=\columnwidth]{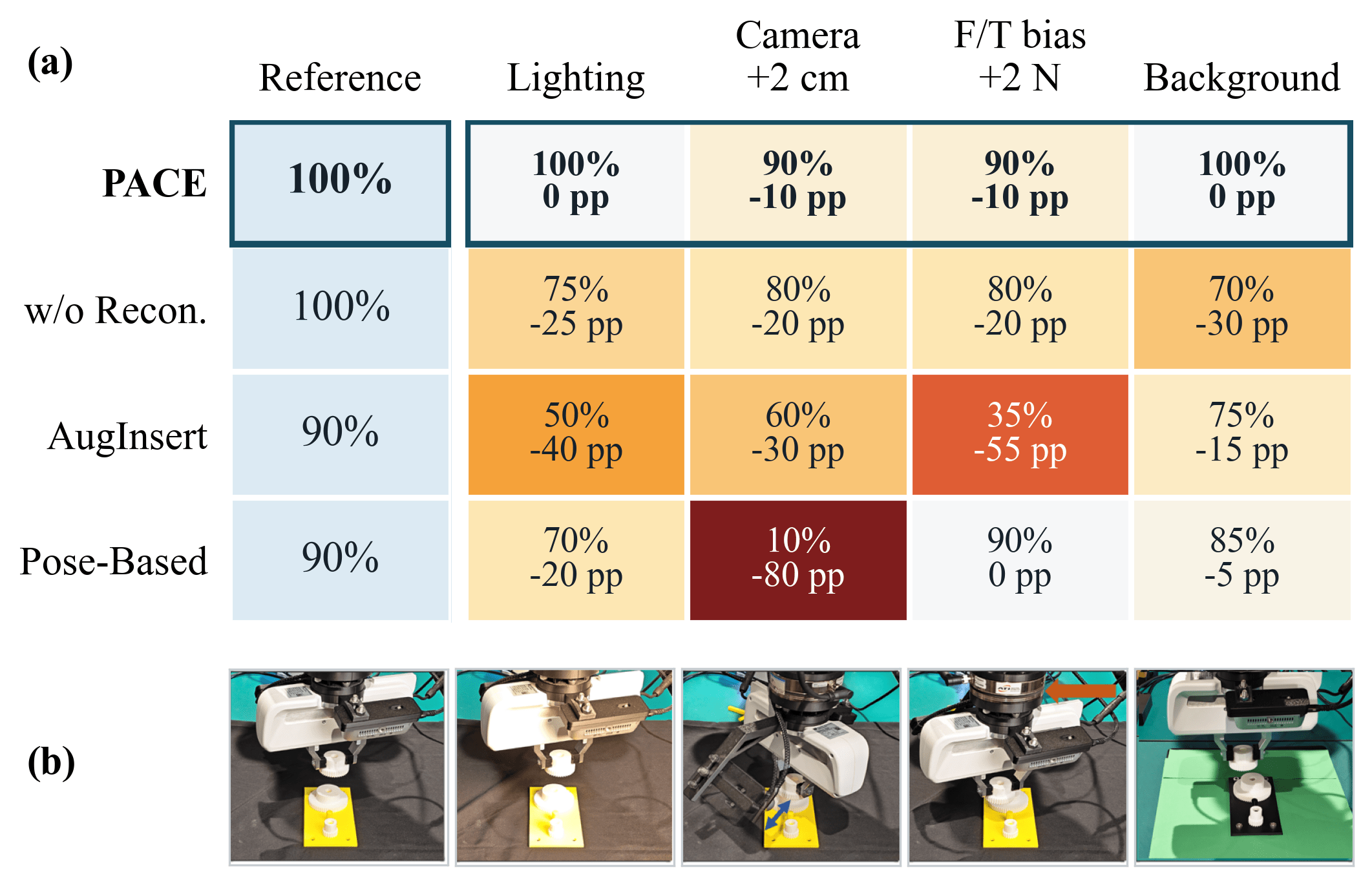}
\caption{Robustness on GearMesh under controlled real-world perturbations. (a) Reference cells report unperturbed success rates; each perturbation cell reports the success rate and its change (pp) from the same method's reference. Darker shading indicates a larger decrease. (b) Reference and perturbed scenes. Each entry is based on 20 hardware trials.}
\label{fig:robustness}
\end{figure}

To test robustness beyond nominal transfer, we evaluate GearMesh under four controlled real-world perturbations: changing lighting, a 2-cm lateral camera shift, a static 2-N F/T bias, and background change. Each method is evaluated in 20 trials under the unperturbed reference condition and each perturbation. Fig.~\ref{fig:robustness} reports success and the change from that method's own reference.

PACE exhibits the smallest degradation: it retains 100\% success under the lighting and background changes and 90\% under both the camera shift and F/T bias, at most 10~pp below its reference. In contrast, Pose-Based PPO drops from 90\% to 10\% under the camera shift, exposing sensitivity to camera displacement. AugInsert drops from 90\% to 50\% under changing lighting and to 35\% under the F/T bias. Without masked reconstruction, performance of PACE decreases by 20--30~pp under every perturbation. Thus, PACE limits degradation across both visual and force perturbations, and the reconstruction ablation indicates that masked reconstruction contributes to this robustness.

\section{Limitations and Future Work}
\label{sec:limitations}

PACE is evaluated on a single robot platform and four quasi-static assembly tasks. Hardware evaluation uses one checkpoint per method--task pair, selected by simulated performance under a common protocol; consequently, the results do not capture transfer variability across training seeds. PACE assumes that calibrated proprioception retains consistent cross-domain semantics, which may weaken with changes in robot embodiment, low-level control, or proprioceptive sensing. This assumption motivates making PACE's proprioceptive anchor more adaptable. Future work could downweight unreliable joint transitions and express motion as end-effector displacement in task space rather than robot-specific joint coordinates. Additional robot-side signals could help maintain a reliable physical reference as sensing conditions change. This approach could support a single pretrained representation across robots, sensing configurations, and longer-horizon tasks.

\section{Conclusion}
\label{sec:conclusion}

We presented PACE, a proprioception-anchored cross-modal encoder for zero-shot sim-to-real transfer in contact-rich assembly. Across four tasks from the Factory and AutoMate benchmarks, policies trained on frozen PACE features transfer to hardware with substantially less performance degradation than policies based on pose estimates or learned multimodal fusion, while remaining robust to visual and sensor perturbations.

The ablations highlight the complementary roles of the main components: proprioceptive anchoring promotes cross-domain generalization, vision supports spatial alignment, F/T sensing resolves contact-sensitive interactions, and masked reconstruction improves robustness when individual observations become unreliable. 
These results suggest a broader principle for sim-to-real representation learning: rather than attempting to eliminate every domain gap, transferable policies can be built by using a physically grounded, domain-stable modality to anchor representations of modalities that are inherently more domain-sensitive.

\bibliographystyle{IEEEtran}
\bibliography{references}

\end{document}